\documentclass[a4paper,titlepage,oneside]{article}
\usepackage{arxiv}

\usepackage{graphicx}
\usepackage{comment}
\usepackage[T1]{fontenc}
\usepackage{mathtools}
\usepackage{amsmath}
\usepackage{amssymb}
\usepackage{float}
\usepackage{slashbox}
\usepackage[ruled,vlined,linesnumbered]{algorithm2e} 

\usepackage{graphicx}
\usepackage[acronym]{glossaries}
\newacronym{iou}{IoU}{Intersection over Union}
\newacronym{sde}{SDE}{stochastic differential equation}
\newacronym{ddpm}{DDPM}{denoising diffusion probabilistic modeling}
\newacronym{mcmc}{MCMC}{Markov chain Monte Carlo}
\newacronym{glas}{GlaS}{Gland Segmentation}
\newacronym{sdf}{SDF}{signed distance function}
\newacronym{gans}{GANs}{generative adversarial networks}
\newacronym{map}{MAP}{maximum a posteriori}
\newacronym{mmse}{MMSE}{minimum mean square error}
\newacronym{cnn}{CNN}{convolutional neural network}
\newacronym{cnns}{CNNs}{convolutional neural networks}
\newacronym{he}{H\&E}{Haematoxylin and Eosin}

\newcommand{\domainimage}{\mathcal{X}}
\newcommand{\domainmask}{\mathcal{M}}
\renewcommand{\d}[1]{\ensuremath{\operatorname{d}\!{#1}}}
\newcommand{\minus}{\scalebox{0.75}[1.0]{$-$}}

\usepackage{siunitx}
\newcommand{\new}{\color{black}}  

\newcommand\blfootnote[1]{%
  \begingroup
  \renewcommand\thefootnote{}\footnote{#1}%
  \addtocounter{footnote}{-1}%
  \endgroup
}

\begin{document}
\title{Score-Based Generative Models for Medical Image Segmentation using Signed Distance Functions}

\author{{Lea Bogensperger}\\
	\texttt{lea.bogensperger@icg.tugraz.at} \\
	\And
	{Dominik Narnhofer} \\
\texttt{dominik.narnhofer@icg.tugraz.at} \\
\And 
{Filip Ilic} \\
\texttt{filip.ilic@icg.tugraz.at} \\
\And 
{Thomas Pock} \\
\texttt{pock@icg.tugraz.at}
}

\renewcommand{\shorttitle}{Score-Based Generative Models for Medical Image Segmentation}

\maketitle              

\begin{abstract}
Medical image segmentation is a crucial task that relies on the ability to accurately identify and isolate regions of interest in medical images. 
Thereby, generative approaches allow to capture the statistical properties of segmentation masks that are dependent on the respective structures.
In this work we propose a conditional score-based generative modeling framework to represent the \gls{sdf} leading to an implicit distribution of segmentation masks.
The advantage of leveraging the \gls{sdf} is a more natural distortion when compared to that of binary masks.
By learning the score function of the conditional distribution of \gls{sdf}s we can accurately sample from the distribution of segmentation masks, allowing for the evaluation of statistical quantities.
Thus, this probabilistic representation allows for the generation of uncertainty maps represented by the variance, which can aid in further analysis and enhance the predictive robustness.
We qualitatively and quantitatively illustrate competitive performance of the proposed method on a public nuclei and gland segmentation data set, highlighting its potential utility in medical image segmentation applications.

\keywords{Score-based generative models  \and image segmentation \and conditional diffusion models \and signed distance function.}
\end{abstract}
\section{Introduction}

Medical image segmentation approaches are often trained end-to-end in a discriminative manner using deep neural networks~\cite{ronneberger2015u,badrinarayanan2017segnet,wang2020axial}. However, also generative models have emerged for image segmentation with the advantage of learning the underlying statistics of segmentation masks conditioned on input images~\cite{iqbal2022generative,xun2022generative,amit2021segdiff}. Apart from \gls{gans}, promising candidates in this field are score-based generative models~\cite{ho2020denoising,song2019generative,song2020score}, which learn the score of a data distribution to sample from the distribution in the framework of a \gls{sde}. Herein, noise is gradually injected to smooth the data distribution until it resembles a simple, tractable prior distribution -- a process which can be reversed with the corresponding time-reverse \gls{sde} relying on the learned score function. \blfootnote{\texttt{https://github.com/leabogensperger/generative-segmentation-sdf}}

Diffusion models can be naturally incorporated to solving inverse problems in medical imaging~\cite{song2021solving}, but they can also be extended to learn a conditional distribution, which makes them well suited for image segmentation. Their potential applicability has already been shown in works~\cite{wu2022medsegdiff,amit2021segdiff,wolleb2022diffusion} using conditional \gls{ddpm}. 
A significant drawback when directly injecting noise on segmentation masks is given by the fact that the distortion process is unnatural with respect to the underlying distribution.
One could argue that the statistics of segmentation masks, which are bimodal or contain very few modes depending on the number semantic classes, are not easy to learn as there is no transition between class modes. A remedy is provided by recalling the \gls{sdf}, a classic image segmentation technique~\cite{osher2004level}, which has also regained attention within newer works on discriminative image segmentation using \gls{cnns}~\cite{naylor2018segmentation,xue2020shape,brissman2021predicting}. 
It is based on the idea that an implicit segmentation map is computed using the \gls{sdf} for which at any given point in the resulting segmentation map the orthogonal distance to the closest boundary point is computed. Additionally, the distance is denoted with a negative sign for interior regions and a positive sign for the background regions. Thus, the \gls{sdf} acts as a shape prior in some sense and it represents a smoother distribution of segmentation masks and thus a smooth transition in class modes. 
Moreover, it naturally promotes smoothness within the transformed binary segmentation map -- which in return is obtained by thresholding the \gls{sdf} map at the object boundaries. 

\begin{figure}[t]
    \centering
    \includegraphics[width=\textwidth]{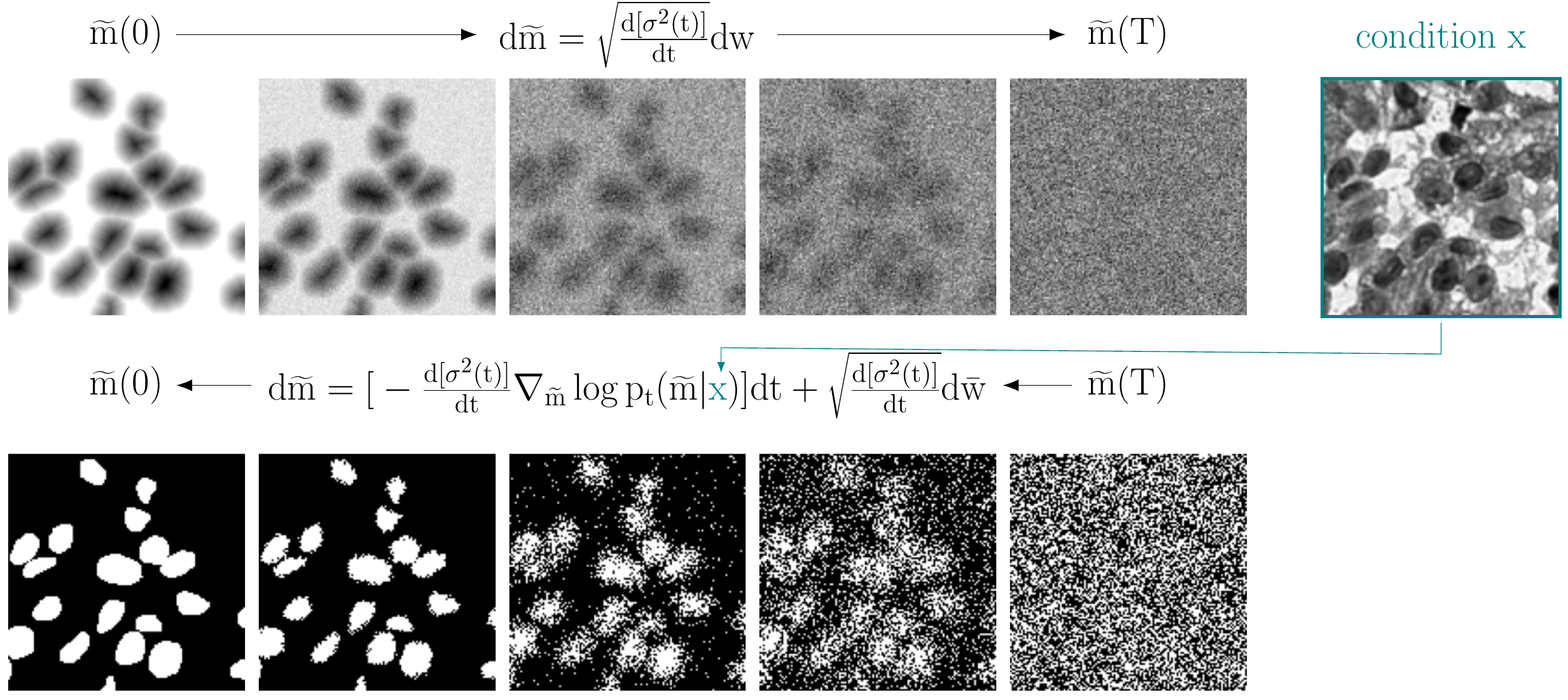}
    \caption{Schematic of the corruption process (top row from left to right for different $t \in [0,T]$) on an \gls{sdf} segmentation mask $\widetilde{m}$ for a given image $x$. The forward and reverse processes are governed by the variance-exploding \gls{sde} and its time-reverse \gls{sde}, respectively, where the latter uses the conditioned score of the distribution of \gls{sdf} masks $\widetilde{m}$ given images $x$. The corresponding thresholded segmentation masks $m$ are shown in the second row.}
    \label{fig:sde}
\end{figure}

In this work, we propose to fuse medical image segmentation using score-based generative modeling based on \gls{sde}s with a segmentation approach relying on the Euclidian \gls{sdf} - thereby learning a smooth implicit representation of segmentation masks conditioned on the respective image. 
The noise-injecting data perturbation process in diffusion models thereby blends in canonically by gradually smoothing the distribution of the \gls{sdf} mask. 
Since binary masks only provide a discrete representation of the object region, the score function can be highly sensitive to small changes in the binary mask, leading to segmentation errors and inconsistencies.
On the other hand, the \gls{sdf} provides a smooth and continuous representation of the object boundary, which can be more natural and better suited to model the shape and boundaries of objects in the image. 


With this approach, the object boundaries are obtained from the sampling process, which can then be used to threshold for the binary segmentation masks, implying that the segmented objects will be smooth. 
Moreover, the segmentation uncertainty can be quantified by acquiring multiple segmentations given an input image due to its generative nature, thus further enhancing the robustness and interpretability of the approach and providing valuable insight into the segmentation process. 

\section{Method}
\subsection{Image Segmentation using \gls{sdf}s}
Image segmentation is the task of finding a segmentation mask $m \in \domainmask$ that assigns each pixel in an image $x \in \domainimage$ a class, where $\domainimage \coloneqq \mathbb{R}^{M\times N}$ and $\domainmask \coloneqq \mathbb{R}^{M \times N}$. Using \gls{sdf}s the segmentation problem can be rephrased in the context of (signed) distances with respect to the object boundaries, where we consider each pixel $m_{ij}$ of the domain $\Omega=\{1,\dots,M\}\times\{1,\dots,N\}$ with the object $\mathcal{S}$ to be segmented. The \gls{sdf} map then contains for each $m_{ij}$ the distance to the closest boundary pixel $\partial \mathcal{S}$, where a negative/positive sign denotes the interior/outside of the object, respectively.

\begin{figure}[t]
    \centering
    \includegraphics[width=\textwidth]{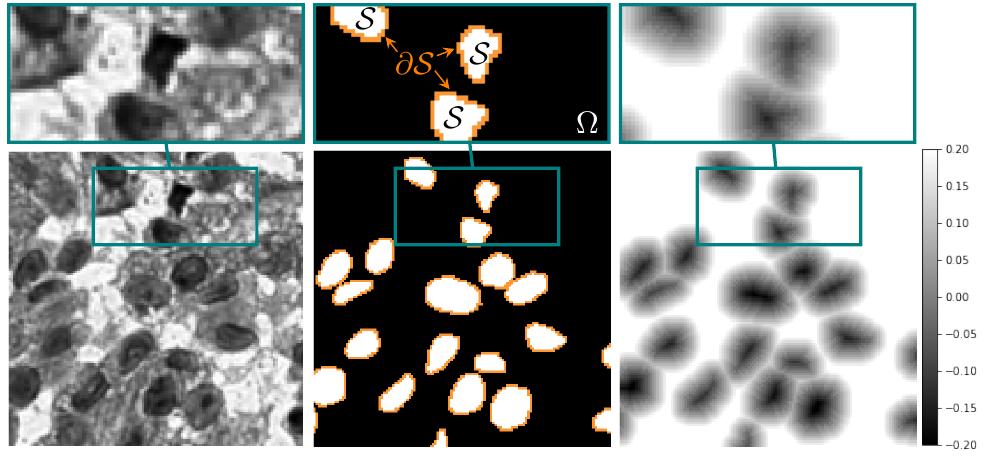}
    \caption{Given an image $x$ (left), its binary segmentation mask (center) can be transformed into a truncated, normalized \gls{sdf} mask (right). The zoomed area shows some segmentation objects in detail denoted by $\mathcal{S}$ and their boundaries $\partial \mathcal{S}$ embedded in the domain $\Omega$.}
    \label{fig:sdf}
\end{figure}

Mathematically, the \gls{sdf} $\widetilde{m}$ of a segmentation mask $m$ can thus be computed using the Euclidian distance function for each pixel $m_{ij}$, which additionally can be truncated at a threshold $\delta$ to consider only a span of pixels around the object boundary $\partial \mathcal{S}$. Moreover, the truncation helps to remain agnostic with respect to higher positive distances belonging to the background, as they should not impact the resulting segmentation. The implicit, truncated \gls{sdf} $\phi(m_{ij})$ is then obtained as follows (also see~\cite{osher2004level}):
\begin{align}
\phi(m_{ij}) = \begin{cases}
 \minus \min \{ \min_{y\in \partial S} \Vert y-m_{ij}\Vert_2, \delta \} \quad &\text{ if } m_{ij} \in \mathcal{S},\\
\min \{ \min_{y\in \partial S} \Vert y-m_{ij}\Vert_2, \delta \} \quad &\text{ if } m_{ij} \in \Omega \setminus \mathcal{S}, \\
0 \quad &\text{ if } m_{ij} \in \partial \mathcal{S}.
\end{cases}
\end{align}
Thus we can obtain the full segmentation map $\widetilde{m}=\phi(m_{ij})_{\substack{i=1,\dots,M \\ j=1,\dots,N}}$ by applying $\phi(\cdot)$ on the entire segmentation mask $m$, which is also depicted in Figure~\ref{fig:sdf}. 
Conversely, given $\widetilde{m}$ one can easily retrieve the binary segmentation map $m$ by thresholding at 0 to separate segmented objects from background. 

\subsection{Conditional Score-Based Segmentation}
In generative approaches for image segmentation
the goal is to sample from the conditional distribution $p(\widetilde{m}|x)$ to obtain a segmentation mask $\widetilde{m}$ given an input image $x$. 
We frame this task in the setting of \gls{sde}s, which gradually corrupt data samples from the data distribution $\widetilde{m}(0)\sim p_0$ until a tractable prior distribution $p_T$ is reached. There exists a corresponding time-reverse \gls{sde} which can then be leveraged to transform the prior distribution back to the data distribution by using the score of the conditional data distribution at time $t$, i.e. $\nabla_{\widetilde{m}} \log p_t(\widetilde{m}|x)$. This score function can be learned using a training set of $S$ paired data samples of images and \gls{sdf} segmentation masks $\mathcal{D} = \{(\widetilde{m}_s,x_s)\}_{s=1}^S$. 

In general, \gls{sde}s have a drift term and a diffusion coefficient that govern its forward and corresponding reverse evolution. Hereby, we solely focus on so-called variance-exploding \gls{sde}s, as presented in~\cite{song2020score}, although our approach should hold also for variance-preserving \gls{sde}s. Thus, for a time process with $t\in[0,T]$, a sequence $\{\widetilde{m}(t)\}_{t=0}^T$ is generated by means of additive corruptive Gaussian noise with standard deviation $\sigma(t)$. Using Brownian motion $w$ to denote the noise corruption, the corresponding \gls{sde} then reads as 
\begin{equation}\label{eq:sde}
    \d{\widetilde{m}} = \sqrt{\frac{\d{[\sigma^2(t)]}}{\d{t}}}\d{w}.
\end{equation}
The \gls{sde} in~\eqref{eq:sde} can be reversed~\cite{anderson1982reverse}, which requires -- when conditioning on images $x$ -- the score of the conditional distribution $\nabla_{\widetilde{m}}\log p_t(\widetilde{m}|x)$. Both the forward and the reverse process are illustratively demonstrated in Figure~\ref{fig:sde}, where the resulting thresholded segmentation masks also show the effect of the corruption process on the explicit binary segmentation mask (which is obtained by thresholding the \gls{sdf} segmentation mask). The reverse \gls{sde} for $\bar{w}$, since time is now going backwards such that $t \in [T,0]$, essentially reads as
\begin{equation} \label{eq:rev_sde}
     \d{\widetilde{m}} = \Big [ -\frac{\d{[\sigma^2(t)]}}{\d{t}} \nabla_{\widetilde{m}} \log p_t(\widetilde{m}|x) \Big ] \d{t} +  \sqrt{\frac{\d{[\sigma^2(t)]}}{\d{t}}}\d{\bar{w}}. 
\end{equation}

The scores of the conditional distribution $\nabla_{\widetilde{m}}\log p_t(\widetilde{m}|x)$ can be estimated using either techniques from score matching~\cite{hyvarinen2005estimation,song2019generative,song2020sliced} or from implicit score estimation such as \gls{ddpm}~\cite{sohl2015deep,ho2020denoising}. Here, we choose to learn a noise-conditional score network $s_{\theta^*}(\widetilde{m}(t),x,\sigma(t))$ for the reverse-time \gls{sde} following the continuous formulation of denoising score matching with $t \in \mathcal{U}(0,T)$ for noise levels from $\sigma_{\text{min}}$ to $\sigma_{\text{max}}$~\cite{song2020score}. Thereby, using Tweedie's formula~\cite{efron2011tweedie} we minimize the following objective:
\begin{equation}
    \theta^* = \arg \min_{\theta} \mathbb{E}_t  \Big \{ \sigma^2(t) \mathbb{E}_{\widetilde{m}(0)} \mathbb{E}_{\widetilde{m}(t)|\widetilde{m}(0)} \big [\Vert s_{\theta}(\widetilde{m}(t),x,\sigma(t)) - \nabla_{\widetilde{m}}\log p_{0t}(\widetilde{m}(t)|\widetilde{m}(0)) \Vert_2^2 \big] \Big\}.
\end{equation}
The perturbation kernel $p_{0t}(\widetilde{m}(t)|\widetilde{m}(0))$ has the form of a standard normal distribution using $\sigma(t)=\sigma_{\min} (\frac{\sigma_{\max}}{\sigma_{\min}})^t$. 

Once the learned scores $s_{\theta^*}$ are available, they can be used to sample from the conditional distribution, where there exist several numerical solvers based on the time-reverse \gls{sde}. Here, we employ a predictor-corrector sampler as proposed by~\cite{song2020score}, which alternates between time-reverse \gls{sde} steps (the predictor) and Langevin \gls{mcmc} sampling steps (the corrector), see Algorithm~\ref{alg:pc_sampling}.
\begin{algorithm}[ht]
\caption{Predictor-corrector algorithm to sample from $p(\widetilde{m}|x)$.}\label{alg:pc_sampling}
\SetAlgoLined 
Choose conditioning image $x$, set number of iterations $K,J$, set $r \in \mathbb{R}^+$ \\
$\widetilde{m}_K \sim \mathcal{N}(0,\sigma_{\max}^2 I)$\\ 
\For{$k = K-1,\ \ldots ,\  0 $}
{
$z \sim \mathcal{N}(0,I)$\;
$\widetilde{m}_k = \widetilde{m}_{k+1} + (\sigma_{k+1}^2 - \sigma_k^2) s_{\theta^*}(\widetilde{m}_{k+1},x,\sigma_{k+1}) + \sqrt{\sigma_{k+1}^2 - \sigma_k^2}z$\; 
\For{$j=1, \ldots, \ J$}
{
$z \sim \mathcal{N}(0,I)$\;
$g =s_{\theta^*}(\widetilde{m}_k^{j-1},x,\sigma_k) $ \;
$\varepsilon = 2(r \Vert z\Vert_2 / \Vert g \Vert_2)^2$ \;
$\widetilde{m}_k^j = \widetilde{m}_k^{j-1} + \varepsilon g + \sqrt{2\varepsilon} z $ \;
}
$\widetilde{m}_{k} = \widetilde{m}_k^J$\;
}
\end{algorithm}

\subsection{Motivation of the \gls{sdf} in Conditional Score-Based Segmentation}

The motivation of using the \gls{sdf} in conditional score-based segmentation is due to the nature of the perturbation process in standard diffusion models, which consists of gradually adding noise to the sought segmentation mask in its \gls{sdf} representation such that its distribution gets smoothed. The destructive process thus implicitly incorporates the boundary information of the segmented objects. 

\begin{figure}[t]
    \centering
    \includegraphics[width=\textwidth]{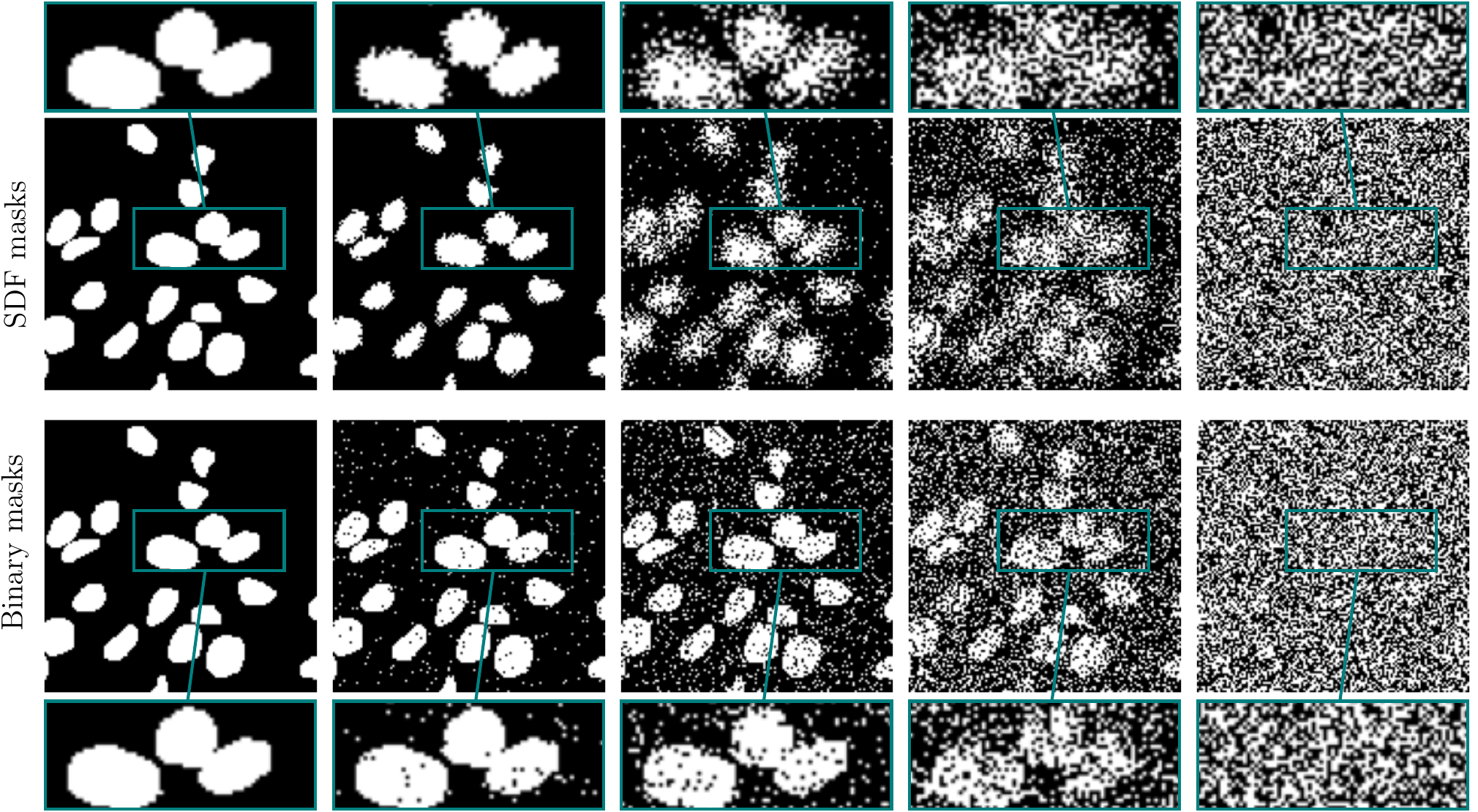}
    \caption{Comparison of the effect of the corruption process on the resulting thresholded segmentation masks when using an \gls{sdf} mask $\widetilde{m}$ and a binary mask $m$ for a given image $x$. Note that the \gls{sdf} representation allows for a more natural distortion process in its thresholded masks, which evolves along the object boundaries instead of directly introducing ``hole''-like structures at random pixel positions as it is the case in the thresholded masks when directly using the binary segmentation mask.}
    \label{fig:sdeNOsdf}
\end{figure}

In contrast, if the binary masks are used directly in the perturbation process, the corruption yields ``hole''-like structures in the resulting thresholded segmentation masks and there is no possibility to integrate the structure of the segmentation objects within the forward process. A comparison of effect of the destructive process on both variants of segmentation mask representations for varied time steps is shown in Figure~\ref{fig:sdeNOsdf}. 

\section{Experiments}
\subsection{Data Sets}
For the experimental setting, we utilize two publicly available data sets. The first data set is MoNuSeg
~\cite{kumar2017dataset}, which consists of 30 training images and 14 test images. Each image is of size $1000\times 1000$ and overall they contain more than 21,000 annotated nuclei in \gls{he} stained microscopic images. For data preprocessing we resort to a structure-preserving color normalization~\cite{vahadane2015structure} as the different organ sites yield considerable intensity variations in the data. This is followed by a gray scale conversion and all images are subsequently resized to $500\times 500$. During training, overlapping crops of $128\times 128$ were used with random horizontal and vertical flips for data augmentation. 

As a second medical data set, the \gls{glas} data set
~\cite{sirinukunwattana2017gland} was used. It consists of 85 training and 80 test \gls{he} stained microscopic images with annotated glands from colorectal cancer tissue. Again, a structure-preserving color normalization~\cite{vahadane2015structure} is used also here for data preprocessing, which is followed by resizing all training and test images to $128\times 128$ inspired by~\cite{valanarasu2021medical}. Data augmentation in training is done as with the MoNuSeg data set.

\subsection{Architecture \& Training}
The architecture to learn the noise-conditional score function is adapted from~\cite{ho2020denoising,song2020score}. 
Further, a conditioning on the image $x$ is required, for which we roughly follow recent works related to conditional generative modeling by concatenating the encoded conditioning image to the network input~\cite{ozdenizci2023restoring,ho2022cascaded,saharia2022palette}. 

For the diffusion parameters we set $\sigma_{\max}=5$ and $\sigma_{\min}=\num{1e-3}$. Note that the latter is slightly lower than usually proposed in literature, which is due to the \gls{sdf} data distribution, as denoising score matching requires a perturbation kernel at the lowest noise scale such that the input distribution remains more or less unchanged. For the learning setting we employ Adam's optimizer~\cite{kingma2014adam} with default coefficient values and a learning rate of \num{1e-4}. 

\subsection{Sampling} To obtain segmentation masks for test images $x$, we use the predictor-corrector sampler in Algorithm~\ref{alg:pc_sampling}. The test images of the MoNuSeg data set are each divided into four evenly sized patches per image, whereas for the \gls{glas} data set the entire test images are processed. In all sampling experiments we set $r=0.35/0.15$ for the corrector step size scaling for the MoNuSeg and \gls{glas} data set, respectively, as we empirically found this to work best in terms of evaluation metrics. Moreover, we use $K=500/200$ predictor steps and $J=2/1$ corrector steps for both data sets, respectively, as this setting revealed to yield best results, despite general low numeric fluctuations amongst different settings. 

\subsection{Evaluation} The resulting samples are \gls{sdf} predictions which have yet to be converted to valid segmentation maps. Thus, they have to be thresholded at $0$ which represents object boundaries to separate segmented objects (which have $0$ at the boundary and negative distances inside assigned) and background (consisting of positive distances). However, due to the employed approach of denoising score matching we set the threshold to $3\sigma_{\min}$, since we have to assume that there is still remaining noise present at the scale of the smallest noise level. This is crucial to consider since we are interested in the exact boundary. 

By leveraging the generative nature of our approach, we are further investigating the effect of averaging over $128$ sampling runs. Thereby we obtain the \gls{mmse} which gives a more robust prediction, which is shown in increased quantitative scores. 

The obtained segmentation masks are then evaluated using the standard metrics F1 score and \gls{iou}. 
We compare our method to commonly referred benchmark models, ensuring that both U-Net variants and attention/transformer mechanisms are considered. Moreover, we also consider~\cite{wolleb2022diffusion} to obtain a comparison with a generative, conditional \gls{ddpm} that predicts standard binary segmentation masks in the sampling process. The approach in~\cite{wolleb2022diffusion} can also be viewed in the form of an \gls{sde} and its time-reverse \gls{sde} using a variance-preserving scheme. 

\subsection{Results}
Table~\ref{tab:results} shows quantitative results for both data sets with our method and comparison methods. To enable a fair comparison the benchmark results were taken from~\cite{valanarasu2021medical} where possible. 
The results indicate that our method outperforms the comparison methods on the \gls{glas} data set, but also for the MoNuSeg data set competitive results can be obtained, although slightly worse than some of the comparison methods. For both data sets, using the \gls{mmse} by averaging over multiple sampling runs clearly gives a significant boost in quantitative performance. 
In comparison, the \gls{ddpm} delivers slightly worse results, but we want to emphasize that they  could also be increased by computing the \gls{mmse} over several runs \new{before thresholding} as it is a generative model -- this was also shown in~\cite{wolleb2022diffusion} where segmentation ensembles are computed to improve the results. 

\begin{table}
\begin{center}
\caption{Quantitative segmentation results on the MoNuSeg and \gls{glas} data set.}\label{tab:results}
\begin{tabular}{p{4.cm}|p{1.4cm} p{1.5cm}|p{1.4cm} p{1.5cm}}
& \multicolumn{2}{c}{MoNuSeg} & \multicolumn{2}{c}{\gls{glas}} \\ \cline{2-5}
\backslashbox[37mm]
{\footnotesize{\textbf{Method}}}{\footnotesize{\textbf{Metric}}}
 &  F1~$\mathrm{\uparrow}$ & m\gls{iou}~$\mathrm{\uparrow}$ &F1~$\mathrm{\uparrow}$ & m\gls{iou}~$\mathrm{\uparrow}$ \\
\hline\hline
FCN~\cite{badrinarayanan2017segnet} & 28.84 &28.71 & 66.61 &50.84 \\
U-Net~\cite{ronneberger2015u} & 79.43 &65.99 & 77.78 & 65.34 \\
U-Net++~\cite{zhou2018unet++} & 79.49& 66.04&78.03 &65.55 \\
Res-UNet~\cite{xiao2018weighted} & 79.49 & 66.07 &78.83 &65.95 \\
Axial Attention U-Net~\cite{wang2020axial} & 76.83 & 62.49 &76.26& 63.03 \\
MedT~\cite{valanarasu2021medical} & \textbf{79.55} & \textbf{66.17} & 81.02& 69.61\\
\gls{ddpm}~\cite{wolleb2022diffusion}& 76.03& 61.42& 76.81&64.15 \\
Ours & 78.13 & 64.19& 82.03 & 71.36  \\
Ours -- \gls{mmse} &78.64 & 64.87 & \textbf{82.77} & \textbf{72.07} \\
\hline
\end{tabular}
\end{center}
\end{table}



\begin{figure}[htb]
    \centering
    \includegraphics[width=\textwidth]{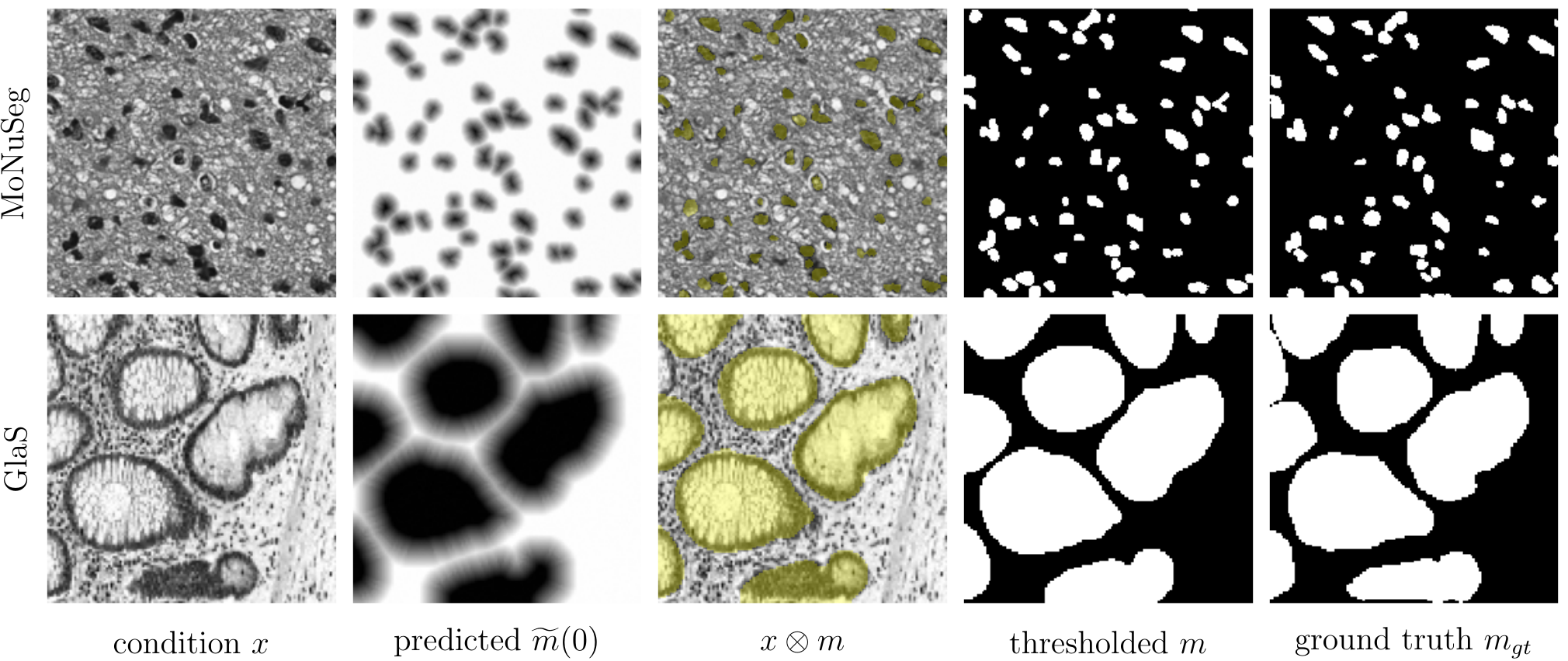}
    \caption{Exemplary sampled segmentation masks for both data sets. The predicted \gls{sdf} masks $\widetilde{m}(0)$ are directly obtained from the sampling procedure, whereas the thresholded masks $m$ are shown to additionally enable a visual comparison with the depicted ground truth $m_{gt}$. Furthermore, we provide the condition image with the overlaid thresholded mask $x\otimes m$.}
    \label{fig:results}
\end{figure}

Figure~\ref{fig:results} shows exemplary qualitative results for both data sets to further highlight the potential applicability of our proposed method. Note that although the evaluation metrics for our approach that can be found in Table~\ref{tab:results} are obtained by evaluating the entire images, we show smaller crops here for the sake of a more detailed visual inspection of the segmented objects. One can clearly observe the smooth segmented objects $m$ obtained from thresholding the \gls{sdf} predictions $\widetilde{m}(0)$, which are in good agreement with the groundtruth segmentation masks $m_{gt}$ for both data set samples.  

\begin{figure}[htb]
    \centering
    \includegraphics[width=\textwidth]{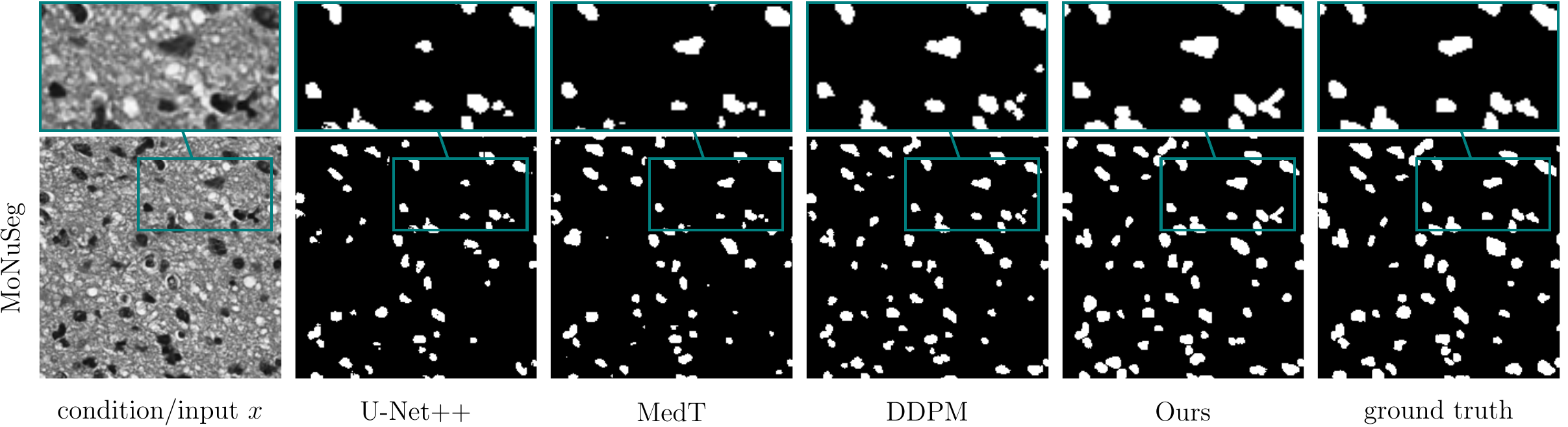}
    \caption{Qualitative results for a test input image $x$ for discriminative approaches resp. condition image $x$ for the generative approaches. We compare our model with the generative \gls{ddpm}~\cite{wolleb2022diffusion} and two discriminative models, U-Net++~\cite{zhou2018unet++} and MedT~\cite{valanarasu2021medical}. Using the \gls{sdf} in our conditional score-based segmentation approach shows that a shape prior is learned, thus preventing small pixel-wise artefacts in the resulting predictions and yielding smooth segmentation objects.}
    \label{fig:results_visual_comp}
\end{figure}

A visual comparison is additionally depicted in Figure~\ref{fig:results_visual_comp}, where we compare the thresholded segmentation prediction of our model of a MoNuSeg test image with its DDPM-based (generative) counterpart~\cite{wolleb2022diffusion} and two discriminative models, namely U-Net++~\cite{zhou2018unet++} and MedT~\cite{valanarasu2021medical}. As can be seen in the provided zoom, the \gls{sdf} representation of segmentation objects indeed seems to act as a shape prior and thus yields smoother segmentation objects while avoiding artefacts such as single pixels/small structures that are mistakenly classified as foreground objects. 

\subsection{Segmentation Uncertainty}

\begin{figure}
    \centering
    \includegraphics[width=\textwidth]{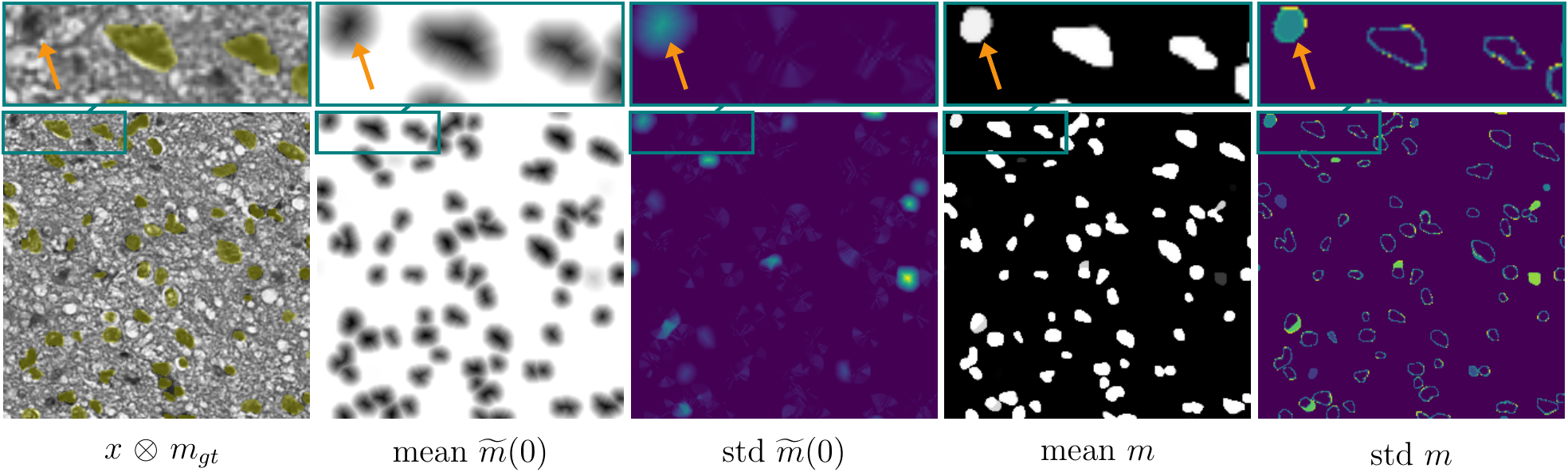}
    \caption{Segmentation example with according statistical values for the \gls{sdf} predictions $\widetilde{m}(0)$ respectively thresholded masks $m$.
    Notably, the image includes a region erroneously segmented as a nuclei, as indicated by the orange arrow. This region is highlighted in the standard deviation maps, which represent the associated uncertainty in the segmentation.
    }
    \label{fig:uncertainty}
\end{figure}

Since the presented approach is based on a generative scheme, we can sample from the conditional distribution of the \gls{sdf} given the conditioning image $p(\widetilde{m}|x)$. 
An advantage of this approach is given by the fact that the resulting statistical values allow for the quantification of segmentation uncertainties in the \gls{sdf} predictions as well as the thresholded masks, which provide additional insights into the segmentation process that are not available with traditional discriminative approaches.
An illustration of the aforementioned property on a MoNuSeg data sample can be seen in Figure~\ref{fig:uncertainty}, where the standard deviation maps associated with the \gls{sdf} predictions and thresholded masks highlight the regions of uncertainty. 

\begin{figure}
    \centering
    \includegraphics[width=\textwidth]{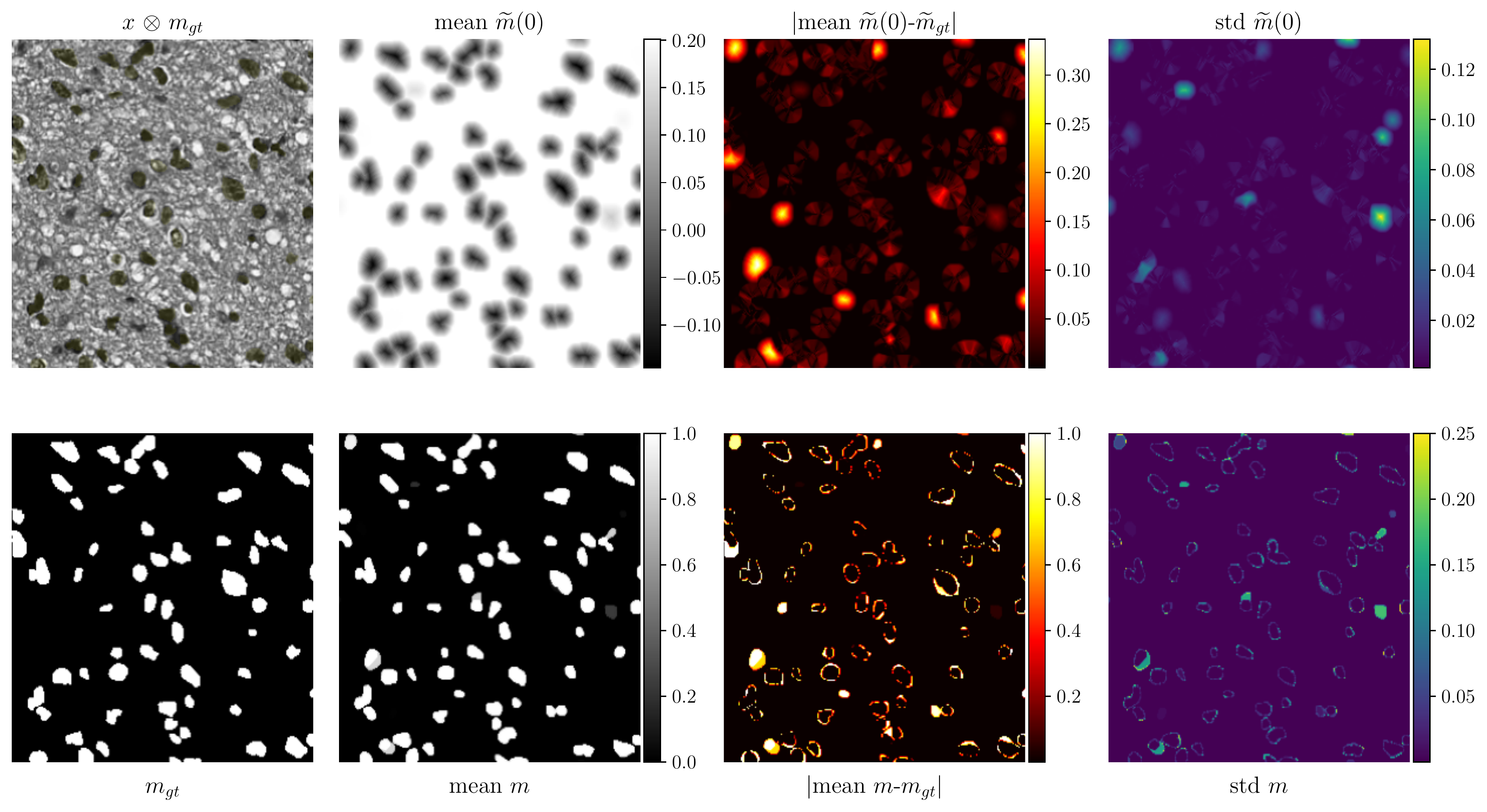}
    \caption{Segmentation example with statistical values for the predicted $\widetilde{m}(0)$ and thresholded masks $m$, along with the corresponding ground truth images and the absolute error between the predictions and the ground truth.
    The visual comparison of error and standard deviation maps (uncertainty) suggests that the latter has predictive capability for the error.}
    \label{fig:uncertainty2}
\end{figure}

In general, we observe that the standard deviation appears high on transitions from nuclei to background as well as in wrongly detected nuclei or over-segmented parts of nuclei. 
This encouraging observation leads us to the hypothesis that the uncertainty may be associated directly with segmentation errors similar to what has been shown in \cite{narnhofer2022posterrquant}, see also Figure~\ref{fig:uncertainty2}. Here, a visual comparison of the error and standard deviation (uncertainty) maps indicates that the latter very likely has high predictive capability.  A detailed analysis including the mutual information of the two variables, however, is out of scope for this work and subject to future work. 

\subsection{Ablation}
As we rely on the standard noise-conditioned score network following~\cite{ho2020denoising} for the proposed method, it is more interesting to investigate the influence of the \gls{sdf}. 
Therefore, we learn the conditional score function such that segmentation masks $m$ can be sampled given $x$, where all other settings remain unchanged -- including $\sigma_{\min}$ and $\sigma_{\max}$.
Hereby, we obtain a mean \gls{iou}/mean F1 score of 51.72/67.80 on the MoNuSeg data set, and 61.39/73.97 on the GlaS data set, respectively. 

Our approach should work reasonably well with other types of architectures suitable for learning the score of a data distribution.
As the primary objective of this work was to introduce a novel concept rather than striving to surpass existing benchmarks, in future work, a more sophisticated network architecture could be considered to further improve the segmentation accuracy.

\section{Discussion and Limitations}
Generative approaches usually require an increased network complexity and are thus computationally more expensive than their discriminative counterparts. We believe that the advantages of having a generative model outweigh its potential drawbacks due to the possibility to evaluate statistical quantities, such as averaging multiple predictions and the generation of uncertainty maps, which may give valuable insights and might be crucial especially in medical imaging applications. \new{While the sampling process itself still requires significantly more time than executing a single forward pass of a discriminative network, there is a variety of new methods available to speed up the inference stage of diffusion models~\cite{song2020denoising,lu2022dpm,karras2022elucidating}. For the sake of simplicity the standard method was chosen, however, there should be nothing to argue against incorporating a different sampling technique.}

The proposed approach outperforms the comparison methods only on one of the two used data sets. However, also on the MoNuSeg data set competitive results have been presented that could potentially be improved with a further optimized network architecture. While a state-of-the art model based on U-Net or transformer architectures might in some cases outperform our quantitative results we still want to emphasize that there is no option of obtaining a segmentation uncertainty \new{as in Figures~\ref{fig:uncertainty} and~\ref{fig:uncertainty2}} with such discriminative approaches. Robustness and reliability should not be traded for minor quantitative increments and improvements. 

Moreover, using \gls{sdf} maps to represent the segmentation masks bears the advantage that a shape prior is learned which favors smooth segmentation objects. The resulting binary segmentation masks thus encompass a different characteristic than when using standard conditional \gls{ddpm} based approaches~\cite{amit2021segdiff,wolleb2022diffusion}.
Although the statistics of these underlying segmentation masks are properly learned, the quantitative results do not reflect those findings in the respective metrics. 


Further, obtaining \gls{sdf} predictions can generally offer other potential advantages such as obtaining instance segmentations as a byproduct. By using the watershed transform, the \gls{sdf} maps could easily be turned into instance segmentation maps and the problem of touching instances is circumvented. 

\section{Conclusion and Outlook}
In this work we proposed a generative approach for medical image segmentation by fusing conditional score-based methods with the concept of representing segmentation maps with the \gls{sdf}. 
The potential applicability of the method was demonstrated qualitatively and quantitatively on two public medical data sets. The \gls{sdf} provides a smoother and more continuous representation of object boundaries compared to binary masks, which makes it a more canonical choice for accurate and robust segmentation.

Furthermore, by leveraging the generative approach, statistical measures such as mean and standard deviation can be calculated to quantify the uncertainty in the segmentation results. 
This information is especially useful for medical diagnosis and treatment planning, where it is important to know the level of confidence in the segmentation results.

As an outlook for future research we will focus on the extension of the approach to multi-class segmentation by additionally incorporating the exclusivity of semantic classes per pixel into the predicted \gls{sdf} maps. 
Additionally, fuelled by preliminary experimental successes, the possibility of learning the score of the joint distribution of images $x$ and segmentation masks $\widetilde{m}$ will be explored. This would provide a powerful framework, where one could directly sample paired training data ${(\widetilde{m}_s,x_s)}$ from the joint distribution or condition on either one of them to sample from the conditional distributions $p(\widetilde{m}|x)$ or $p(x|\widetilde{m})$.



\newpage

\bibliographystyle{splncs04}
\bibliography{main}

\end{document}